\documentclass[a4paper]{article}
\usepackage{amsfonts,amsmath,amssymb,amsthm,graphicx,color,algorithmic,hyperref}
\usepackage{RR}

\providecommand{\U}[1]{\protect\rule{.1in}{.1in}}
 \DeclareMathOperator{\tr}{tr}

 \DeclareMathOperator{\diag}{diag}
\DeclareMathOperator{\Diag}{Diag}

\RRdate{December 2009} %%

\RRauthor{% les auteurs
Ngoc Q.K. Duong\thanks{qduong@irisa.fr}%
  \and
Emmanuel Vincent\thanks{emmanuel.vincent@inria.fr}%
\and
R\'emi Gribonval\thanks{remi.gribonval@inria.fr} }
\authorhead{Duong, Vincent, and Gribonval}

\RRtitle{S\'eparation de m\'elanges audio r\'everb\'erants
sous-d\'eterminés à l'aide d'un mod\`ele de covariance spatiale de
rang plein} %%

\RRetitle{Under-determined reverberant audio source separation using
a full-rank spatial covariance model} %%
\titlehead{Under-determined reverberant audio source separation}

\RRresume{Cet article traite de la mod\'elisation d'environnements
d'enregistrement r\'everb\'erants dans le contexte de la
s\'eparation de sources sous-d\'etermin\'ee. Nous mod\'elisons la
contribution de chaque source à l'ensemble des canaux du m\'elange
dans le domaine temps-fr\'equence comme une variable al\'eatoire
vectorielle gaussienne de moyenne nulle dont la covariance code les
caract\'eristiques spatiales de la source. Nous consid\'erons quatre
mod\`eles sp\'ecifiques de covariance, dont un mod\`ele de rang
plein non contraint. Nous explicitons une famille d'algorithmes
Expectation-Maximization (EM) pour l'estimation des param\`etres de
chaque mod\`ele et nous proposons des proc\'edures ad\'equates
d'initialisation des param\`etres et d'appariement de l'ordre des
sources à travers les fr\'equences à partir de leurs directions
d'arriv\'ee. Les r\'esultats exp\'erimentaux sur des m\'elanges
r\'everb\'erants synth\'etiques et enregistr\'es montrent la
pertinence de l'approche propos\'ee.}

\RRabstract{This article addresses the modeling of reverberant
recording environments in the context of under-determined
convolutive blind source separation. We model the contribution of
each source to all mixture channels in the time-frequency domain as
a zero-mean Gaussian random variable whose covariance encodes the
spatial characteristics of the source. We then consider four
specific covariance models, including a full-rank unconstrained
model. We derive a family of iterative expectation-maximization (EM)
algorithms to estimate the parameters of each model and propose
suitable procedures to initialize the parameters and to align the
order of the estimated sources across all frequency bins based on
their estimated directions of arrival (DOA). Experimental results
over reverberant synthetic mixtures and live recordings of speech
data show the effectiveness of the proposed approach.}

\RRmotcle{S\'eparation de sources convolutive, m\'elanges
sous-d\'etermin\'es, mod\`eles de covariance spatiale, algorithme
EM, probl\`eme de permutation.}

\RRkeyword{Convolutive blind source separation, under-determined
mixtures, spatial covariance models, EM algorithm, permutation
problem.}

\RRprojet{METISS} \RRtheme{\THCom}
\RCRennes % Sophia Antipolis M\'editerran\'ee

\begin{document}
\RRNo{7116}
\makeRR   % cas d'un rapport de recherche

\section{Introduction}
\label{sec:intro}

In blind source separation (BSS), audio signals are generally
mixtures of several sound sources such as speech, music, and
background noise. The recorded multichannel signal $\mathbf{x}(t)$
is therefore expressed as
\begin{equation}
\mathbf{x}(t)=\sum_{j=1}^J \mathbf{c}_j(t) \label{eq:origmix}
\end{equation}
where $\mathbf{c}_j(t)$ is the spatial image of the $j$th source,
that is the contribution of this source to all mixture channels. For
a point source in a reverberant environment, $\mathbf{c}_j(t)$ can
be expressed via the convolutive mixing process
\begin{equation}
\mathbf{c}_j(t)=\sum_\tau \mathbf{h}_j(\tau) s_j(t-\tau)
\label{eq:conv}
\end{equation}
where $s_j(t)$ is the $j$th source signal and $\mathbf{h}_j(\tau)$
the vector of filter coefficients modeling the acoustic path from
this source to all microphones. Source separation consists in
recovering either the $J$ original source signals or their spatial
images given the $I$ mixture channels. In the following, we focus on
the separation of under-determined mixtures, \textit{i.e.} such that
$I<J$.

Most existing approaches operate in the time-frequency domain using
the short-time Fourier transform (STFT) and rely on narrowband
approximation of the convolutive mixture \eqref{eq:conv} by
complex-valued multiplication in each frequency bin $f$ and time
frame $n$ as
\begin{equation}
\mathbf{c}_j(n,f)\approx \mathbf{h}_j(f)s_j(n,f) \label{eq:insmix}
\end{equation}
where the mixing vector $\mathbf{h}_j(f)$ is the Fourier transform
of $\mathbf{h}_j(\tau)$, $s_j(n,f)$ are the STFT coefficients of the
sources $s_j(t)$ and $\mathbf{c}_j(n,f)$ the STFT coefficients of
their spatial images $\mathbf{c}_j(t)$. The sources are typically
estimated under the assumption that they are sparse in the STFT
domain. For instance, the degenerate unmixing estimation technique
(DUET) \cite{O.Yilmaz-IEEETrans04} uses binary masking to extract
the predominant source in each time-frequency bin. Another popular
technique known as $\ell_1$-norm minimization extracts on the order
of $I$ sources per time-frequency bin by solving a constrained
$\ell_1$-minimization problem \cite{Winter07,Bofill}. The separation
performance achievable by these techniques remains limited in
reverberant environments \cite{E.Vincent-SiSEC08}, due in particular
to the fact that the narrowband approximation does not hold because
the mixing filters are much longer than the window length of the
STFT.

Recently, a distinct framework has emerged whereby the STFT
coefficients of the source images $\mathbf{c}_j(n,f)$ are modeled by
a phase-invariant multivariate distribution whose parameters are
functions of $(n,f)$ \cite{VincentIGI}. One instance of this
framework consists in modeling $\mathbf{c}_j(n,f)$ as a zero-mean
Gaussian random variable with covariance matrix
\begin{equation}
\mathbf{R}_{\mathbf{c}_j}(n,f)=v_j(n,f)\,\mathbf{R}_j(f)
\label{eq:Cov}
\end{equation}
where $v_j(n,f)$ are scalar time-varying \emph{variances} encoding
the spectro-temporal power of the sources and $\mathbf{R}_j(f)$ are
time-invariant \emph{spatial covariance matrices} encoding their
spatial position and spatial spread \cite{Duong-WASPAA09}. The model
parameters can then be estimated in the maximum likelihood (ML)
sense and used estimate the spatial images of all sources by Wiener
filtering.

This framework was first applied to the separation of instantaneous
audio mixtures in \cite{Fevotte-WASPAA05, E.Vincent-ICA09} and shown
to provide better separation performance than $\ell_1$-norm
minimization. The instantaneous mixing process then translated into
a rank-1 spatial covariance matrix for each source. In our
preliminary paper \cite{Duong-WASPAA09}, we extended this approach
to convolutive mixtures and proposed to consider full-rank spatial
covariance matrices modeling the spatial spread of the sources and
circumventing the narrowband approximation. This approach was shown
to improve separation performance of reverberant mixtures in both an
\emph{oracle} context, where all model parameters are known, and in
a \emph{semi-blind} context, where the spatial covariance matrices
of all sources are known but their variances are blindly estimated
from the mixture.

In this article we extend this work to \emph{blind} estimation of
the model parameters for BSS application. While the general
expectation-maximization (EM) algorithm is well-known as an
appropriate choice for parameter estimation of Gaussian models
\cite{EMref,Cardoso-eusipco02,S.Arberet-ICA09,Izumi-WASPAA07}, it is
very sensitive to the initialization \cite{Ozerov-IEEETrans10}, so
that an effective parameter initialization scheme is necessary.
Moreover, the well-known source permutation problem arises when the
model parameters are independently estimated at different
frequencies \cite{H.Sawada-IEEETrans07}. In the following, we
address these two issues for the proposed models and evaluate these
models together with state-of-the-art techniques on a considerably
larger set of mixtures.

The structure of the rest of the article is as follows. We introduce
the general framework under study as well as four specific spatial
covariance models in Section \ref{sec:models}. We then address the
blind estimation of all model parameters from the observed mixture
in Section \ref{sec:blind}. We compare the source separation
performance achieved by each model to that of state-of-the-art
techniques in various experimental settings in Section
\ref{sec:experiment}. Finally we conclude and discuss further
research directions in Section \ref{sec:conclusion}.

\section{General framework and spatial covariance models}
\label{sec:models} We start by describing the general probabilistic
modeling framework adopted from now on. We then define four models
with different degrees of flexibility resulting in rank-1 or
full-rank spatial covariance matrices.

\subsection{General framework}
Let us assume that the vector $\mathbf{c}_j(n,f)$ of STFT
coefficients of the spatial image of the $j$th source follows a
zero-mean Gaussian distribution whose covariance matrix factors as
in \eqref{eq:Cov}. Under the classical assumption that the sources
are uncorrelated, the vector $\mathbf{x}(n,f)$ of STFT coefficients
of the mixture signal is also zero-mean Gaussian with covariance
matrix
\begin{equation}
\mathbf{R}_{\mathbf{x}}(n,f)=\sum_{j=1}^J v_j(n,f)\,\mathbf{R}_j(f).
\label{eq:Rx}
\end{equation}
In other words, the likelihood of the set of observed mixture STFT
coefficients $\mathbf{x}=\{\mathbf{x}(n,f)\}_{n,f}$ given the set of
variance parameters $v=\{v_j(n,f)\}_{j,n,f}$ and that of spatial
covariance matrices $\mathbf{R}=\{\mathbf{R}_j(f)\}_{j,f}$ is given
by
\begin{equation}
P(\mathbf{x}|v,\mathbf{R})=\prod_{n,f}\frac{1}{\det\left(\pi\mathbf{R}_{\mathbf{x}}(n,f)\right)}e^{-\mathbf{x}^H(n,f)\mathbf{R}_\mathbf{x}^{-1}(n,f)\mathbf{x}(n,f)}
\end{equation}
where $^H$ denotes matrix conjugate transposition and
$\mathbf{R}_{\mathbf{x}}(n,f)$ implicitly depends on $v$ and
$\mathbf{R}$ according to \eqref{eq:Rx}. The covariance matrices are
typically modeled by higher-level spatial parameters, as we shall
see in the following.

Under this model, source separation can be achieved in two steps.
The variance parameters $v$ and the spatial parameters underlying
$\mathbf{R}$ are first estimated in the ML sense. The spatial images
of all sources are then obtained in the minimum mean square error
(MMSE) sense by multichannel Wiener filtering
\begin{equation}
\widehat{\mathbf{c}}_j(n,f)=v_j(n,f)\mathbf{R}_j(f)\mathbf{R}_\mathbf{x}^{-1}(n,f)\mathbf{x}(n,f).
\label{eq:Wienerfiltering}
\end{equation}

\subsection{Rank-$1$ convolutive model}
\label{ssec:model1_conv} Most existing approaches to audio source
separation rely on narrowband approximation of the convolutive
mixing process \eqref{eq:conv} by the complex-valued multiplication
\eqref{eq:insmix}. The covariance matrix of $\mathbf{c}_j(n,f)$ is
then given by \eqref{eq:Cov} where $v_j(n,f)$ is the variance of
$s_j(n,f)$ and $\mathbf{R}_j(f)$ is equal to the rank-1 matrix
\begin{equation}
\mathbf{R}_j(f)=\mathbf{h}_j(f)\mathbf{h}_j^H(f) \label{eq:Model1}
\end{equation}
with $\mathbf{h}_j(f)$ denoting the Fourier transform of the mixing
filters $\mathbf{h}_j(\tau)$. This \emph{rank-1 convolutive model}
of the spatial covariance matrices has recently been exploited in
\cite{Ozerov-IEEETrans10} together with a different model of the
source variances.

\subsection{Rank-$1$ anechoic model}
\label{ssec:model1_ane}

In an anechoic recording environment without reverberation, each
mixing filter boils down to the combination of a delay $\tau_{ij}$
and a gain $\kappa_{ij}$ specified by the distance $r_{ij}$ from the
$j$th source to the $i$th microphone \cite{T.Gustafsson-IEEETrans03}
\begin{equation}
\tau_{ij}=\frac{r_{ij}}{c} \label{eq:intensity} \quad\text{and}\quad
\kappa_{ij}=\frac{1}{\sqrt{4\pi} r_{ij}}
\end{equation}
where $c$ is sound velocity. The spatial covariance matrix of the
$j$th source is hence given by the \emph{rank-1 anechoic model}
\begin{equation}
\mathbf{R}_j(f)=\mathbf{a}_j(f)\mathbf{a}_j^H(f)
\label{eq:Model1_ane}
\end{equation}
where the Fourier transform $\mathbf{a}_j(f)$ of the mixing filters
is now parameterized as
\begin{equation}
\mathbf{a}_j(f)=
\begin{pmatrix}
\kappa_{1,j}e^{-2i\pi f\tau _{1,j}}\\
\vdots\\
\kappa_{I,j}e^{-2i\pi f\tau _{I,j}}
\end{pmatrix}. \label{eq:aj}
\end{equation}

\subsection{Full-rank direct+diffuse model}
\label{ssec:model2} One possible interpretation of the narrowband
approximation is that the sound of each source as recorded on the
microphones comes from a single spatial position at each frequency
$f$, as specified by $\mathbf{h}_j(f)$ or $\mathbf{a}_j(f)$. This
approximation is not valid in a reverberant environment, since
reverberation induces some spatial spread of each source, due to
echoes at many different positions on the walls of the recording
room. This spread translates into full-rank spatial covariance
matrices.

The theory of statistical room acoustics assumes that the spatial
image of each source is composed of two uncorrelated parts: a direct
part modeled by $\mathbf{a}_j(f)$ in \eqref{eq:aj} and a reverberant
part. The spatial covariance $\mathbf{R}_j(f)$ of each source is
then a full-rank matrix defined as the sum of the covariance of its
direct part and the covariance of its reverberant part such that
\begin{equation}
\mathbf{R}_j(f)=\mathbf{a}_j(f)\mathbf{a}_j^H(f)+\sigma_\mathrm{rev}^2\mathbf{\Psi}(f)\label{eq:Model2}
\end{equation}
where $\sigma_\mathrm{rev}^2$ is the variance of the reverberant
part and $\Psi_{il}(f)$ is a function of the distance $d_{il}$
between the $i$th and the $l$th microphone such that
$\Psi_{ii}(f)=1$. This model assumes that the reverberation recorded
at all microphones has the same power but is correlated as
characterized by
 $\Psi(d_{il},f)$. This model has been employed for single source localization in \cite{T.Gustafsson-IEEETrans03} but not for source separation yet.

Assuming that the reverberant part is diffuse, \textit{i.e.} its
intensity is uniformly distributed over all possible directions, its
normalized cross-correlation can be shown to be real-valued and
equal to \cite{H.Kuttruff}
\begin{equation}
\Psi_{il}(f)=\frac{\sin(2\pi fd_{il}/c)}{2\pi fd_{il}/c}.
\label{eq:sincfun}
\end{equation}
Moreover, the power of the reverberant part within a
parallelepipedic room with dimensions $L_x$, $L_y$, $L_z$ is given
by
\begin{equation}
\sigma_\mathrm{rev}^{2}=\frac{4\beta ^{2}}{\mathcal{A}(1-\beta^{2})}
\label{eq:sigma}
\end{equation}
where $\mathcal{A}$ is the total wall area and $\beta$ the wall
reflection coefficient computed from the room reverberation time
$T_{60}$ via Eyring's formula \cite{T.Gustafsson-IEEETrans03}
\begin{equation}
\beta=\exp\bigg\{-\frac{13.82}{(\frac{1}{L_x}+\frac{1}{L_y}+\frac{1}{L_z})cT_{60}}\bigg\}.
\end{equation}

\subsection{Full-rank unconstrained model}
\label{ssec:model3} In practice, the assumption that the reverberant
part is diffuse is rarely satisfied. Indeed, early echoes containing
more energy are not uniformly distributed on the walls of the
recording room, but at certain positions depending on the position
of the source and the microphones. When performing some simulations
in a rectangular room, we observed that \eqref{eq:sincfun} is valid
on average when considering a large number of sources at different
positions, but generally not valid for each source considered
independently.

Therefore, we also investigate the modeling of each source via an
unconstrained spatial covariance matrix $\mathbf{R}_j(f)$ whose
coefficients are not related a priori. Since this model is more
general than $\eqref{eq:Model1}$ and $\eqref{eq:Model2}$, it allows
more flexible modeling of the mixing process and hence potentially
improves separation performance of real-world convolutive mixtures.

\section{Blind estimation of the model parameters}
\label{sec:blind} 
\begin{figure}[htb]
\centering
\includegraphics[scale=0.75]{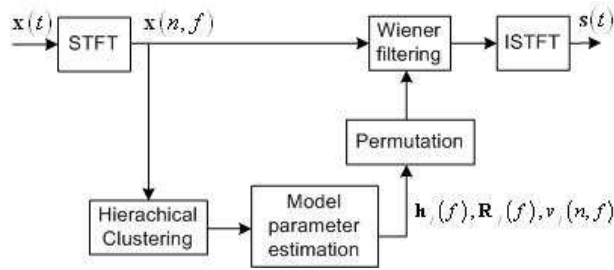}
\caption{Flow of the proposed blind source separation approach.}
\label{fig:framework}
\end{figure}

In order to use the above models for BSS, we now
need to estimate their parameters from the observed mixture signal
only. In our preliminary paper \cite{Duong-WASPAA09}, we used a
quasi-Newton algorithm for semi-blind separation that converged in a
very small number of iterations. However, due to the complexity of
each iteration, we later found out that the EM algorithm provided
faster convergence in practice despite a larger number of
iterations. We hence choose EM for blind separation in the
following. More precisely, we adopt the following three-step
procedure: initialization of $\mathbf{h}_j(f)$ or $\mathbf{R}_j(f)$
by hierarchical clustering, iterative ML estimation of all model
parameters via EM, and permutation alignment. The latter step is
needed only for the rank-1 convolutive model and the full-rank
unconstrained model whose parameters are estimated independently in
each frequency bin. The overall procedure is depicted in Fig.
\ref{fig:framework}.

\subsection{Initialization by hierarchical clustering}
\label{ssec:clustering} Preliminary experiments showed that the
initialization of the model parameters greatly affects the
separation performance resulting from the EM algorithm. In the
following, we propose a hierarchical clustering-based initialization
scheme inspired from the algorithm in \cite{Winter07}.

This scheme relies on the assumption that the sound from each source
comes from a certain region of space at each frequency $f$, which is
different for all sources. The vectors $\mathbf{x}(n,f)$ of mixture
STFT coefficients are then likely to cluster around the direction of
the associated mixing vector $\mathbf{h}_j(f)$ in the time frames
$n$ where the $j$th source is predominant.

In order to estimate these clusters, we first normalize the vectors
of mixture STFT coefficients as
\begin{equation}
{\mathbf{\bar x}}(n,f) \leftarrow \frac{\mathbf{x}(n,f)} {\|
\mathbf{x}(n,f) \|_2 } e^{ -i \arg (x_1 (n,f))} \label{eq:normalize}
\end{equation}
where $\arg(.)$ denotes the phase of a complex number and $\|.\|_2$
the Euclidean norm. We then define the distance between two clusters
$C_1$ and $C_2$ by the average distance between the associated
normalized mixture STFT coefficients
\begin{equation}
d(C_1,C_2)=\frac{1}{|C_1||C_2|}\sum_{\mathbf{\bar x}_1 \in C_1}
\sum_{\mathbf{\bar x}_2 \in C_2} \| \mathbf{\bar x}_1 - \mathbf{\bar
x}_2 \|_2 \label{eq:clusterdistance}
\end{equation}

In a given frequency bin, the vectors of mixture STFT coefficients
on all time frames are first considered as clusters containing a
single item. The distance between each pair of clusters is computed
and the two clusters with the smallest distance are merged. This
"bottom up" process called linking is repeated until the number of
clusters is smaller than a predetermined threshold $K$. This
threshold is usually much larger than the number of sources $J$
\cite{Winter07}, so as to eliminate outliers. We finally choose the
$J$ clusters with the largest number of samples. The initial mixing
vector and spatial covariance matrix for each source are then
computed as
\begin{align}
\mathbf{h}_j^\mathrm{init} (f)&=\frac {1}{|C_j|}
\sum_{\mathbf{\bar{x}}(n,f) \in C_j}
\mathbf{\tilde{x}}(n,f) \label{eq:h} \\
\mathbf{R}_j^\mathrm{init} (f)&=\frac {1}{|C_j|}
\sum_{\mathbf{\bar{x}}(n,f) \in C_j} \mathbf{\tilde{x}}(n,f)
\mathbf{\tilde{x}}(n,f)^H
\end{align}
where $\mathbf{\tilde{x}}(n,f)=\mathbf{x}(n,f)e^{-i\arg(x_1(n,f))}$.
Note that, contrary to the algorithm in \cite{Winter07}, we define
the distance between clusters as the average distance between the
normalized mixture STFT coefficients instead of the minimum distance
between them. Besides, the mixing vector $\mathbf{h}_j^\mathrm{init}
(f)$ is computed from the phase-normalized mixture STFT coefficients
$\mathbf{\tilde{x}}(n,f)$ instead of both phase and amplitute
normalized coefficients $\mathbf{\bar{x}}(n,f)$. These modifications
were found to provide better initial approximation of the mixing
parameters in our experiments. We also tested random initialization
and direction-of-arrival (DOA) based initialization, \textit{i.e.}
where the mixing vectors $\mathbf{h}_j^\mathrm{init}(f)$ are derived
from known source and microphone positions assuming no
reverberation. Both schemes were found to result in slower
convergence and poorer separation performance than the proposed
scheme.

\subsection{EM updates for the rank-1 convolutive model}
The derivation of the EM parameter estimation algorithm for the
rank-1 convolutive model is strongly inspired from the study in
\cite{Ozerov-IEEETrans10}, which relies on the same model of spatial
covariance matrices but on a distinct model of source variances.
Similarly to \cite{Ozerov-IEEETrans10}, EM cannot be directly
applied to the mixture model \eqref{eq:origmix} since the estimated
mixing vectors remain fixed to their initial value. This issue can
be addressed by considering the noisy mixture model
\begin{equation}
\mathbf{x}(n,f)=\mathbf{H}(f)\mathbf{s}(n,f)+\mathbf{b}(n,f)
\label{eq:rank1model}
\end{equation}
where $\mathbf{H}(f)$ is the mixing matrix whose $j$th column is the
mixing vector $\mathbf{h}_j(f)$, $\mathbf{s}(n,f)$ is the vector of
source STFT coefficients $s_j(n,f)$ and $\mathbf{b}(n,f)$ some
additive zero-mean Gaussian noise. We denote by
$\mathbf{R}_\mathbf{s}(n,f)$ the diagonal covariance matrix of
$\mathbf{s}(n,f)$. Following \cite{Ozerov-IEEETrans10}, we assume
that $\mathbf{b}(n,f)$ is stationary and spatially uncorrelated and
denote by $\mathbf{R}_\mathbf{b}(f)$ its time-invariant diagonal
covariance matrix. This matrix is initialized to a small value
related to the average accuracy of the mixing vector initialization
procedure.

EM is separately derived for each frequency bin $f$ for the
\emph{complete data} $\{\mathbf{x}(n,f), s_j(n,f)\}_{j,n}$ that is
the set of mixture and source STFT coefficients of all time frames.
The details of one iteration are as follows. In the E-step, the
Wiener filter $\mathbf{W}(n,f)$ and the conditional mean
$\widehat{\mathbf{s}}(n,f)$ and covariance
$\widehat{\mathbf{R}}_{\mathbf{ss}}(n,f)$ of the sources are
computed as
\begin{align}
\mathbf{R}_\mathbf{s}(n,f)&=\diag(v_1(n,f),...,v_J(n,f))\\
\mathbf{R}_\mathbf{x}(n,f)&=\mathbf{H}(f)\mathbf{R}_\mathbf{s}(n,f)\mathbf{H}^H(f)+\mathbf{R}_\mathbf{b}(f)\\
\mathbf{W}(n,f)&=\mathbf{R}_\mathbf{s}(n,f)\mathbf{H}^H(f)\mathbf{R}^{-1}_\mathbf{x}(n,f)\\
\widehat{\mathbf{s}}(n,f)&=\mathbf{W}(n,f)\mathbf{x}(n,f)\\
\widehat{\mathbf{R}}_\mathbf{ss}(n,f)&=\widehat{\mathbf{s}}(n,f)\widehat{\mathbf{s}}^H(n,f)+(\mathbf{I}-\mathbf{W}(n,f)\mathbf{H}(f))\mathbf{R}_\mathbf{s}(n,f)
\end{align}
where $\mathbf{I}$ is the $I\times I$ identity matrix and $\diag(.)$
the diagonal matrix whose entries are given by its arguments.
Conditional expectations of multichannel statistics are also
computed by averaging over all $N$ time frames as
\begin{align}
\widehat{\mathbf{R}}_\mathbf{ss}(f)&=\frac{1}{N}\sum_{n=1}^{N}\widehat{\mathbf{R}}_\mathbf{ss}(n,f)\\
\widehat{\mathbf{R}}_\mathbf{xs}(f)&=\frac{1}{N}\sum_{n=1}^{N}\mathbf{x}(n,f)\widehat{\mathbf{s}}^H(n,f)\\
\widehat{\mathbf{R}}_\mathbf{xx}(f)&=\frac{1}{N}\sum_{n=1}^{N}\mathbf{x}(n,f)\mathbf{x}^H(n,f).
\end{align}
In the M-step, the source variances, the mixing matrix and the noise
covariance are updated via
\begin{align}
v_j(n,f)=&\widehat{\mathbf{R}}_ {\mathbf{ss}\,jj}(n,f)\\
\mathbf{H}(f)=&\widehat{\mathbf{R}}_\mathbf{xs}(f)\widehat{\mathbf{R}}^{-1}_\mathbf{ss}(f)\\
\mathbf{R}_\mathbf{b}(f)=&\Diag(\widehat{\mathbf{R}}_\mathbf{xx}(f)-
\mathbf{H}(f)\widehat{\mathbf{R}}^H_\mathbf{xs}(f)\notag\\
&-\widehat{\mathbf{R}}_\mathbf{xs}\mathbf{H}^H(f)
+\mathbf{H}(f)\widehat{\mathbf{R}}_\mathbf{ss}(n,f)\mathbf{H}^H(f))
\end{align}
where $\Diag(.)$ projects a matrix onto its diagonal.

\subsection{EM updates for the full-rank unconstrained model}
The derivation of EM for the full-rank unconstrained model is much
easier since the above issue does not arise. We hence stick with the
exact mixture model \eqref{eq:origmix}, which can be seen as an
advantage of full-rank \textit{vs.} rank-1 models. EM is again
separately derived for each frequency bin $f$. Since the mixture can
be recovered from the spatial images of all sources, the complete
data reduces to $\{\mathbf{c}_j(n,f)\}_{n,f}$, that is the set of
STFT coefficients of the spatial images of all sources on all time
frames. The details of one iteration are as follows. In the E-step,
the Wiener filter $\mathbf{W}_j(n,f)$ and the conditional mean
$\widehat{\mathbf{c}}_j(n,f)$ and covariance
$\widehat{\mathbf{R}}_{\mathbf{c}_j}(n,f)$ of the spatial image of
the $j$th source are computed as
\begin{align}
\mathbf{W}_j(n,f)&=\mathbf{R}_{\mathbf{c}_j}(n,f)\mathbf{R}_{\mathbf{x}}^{-1}(n,f)\\
\widehat{\mathbf{c}}_j(n,f)&=\mathbf{W}_j(n,f)\mathbf{x}(n,f)\\
\widehat{\mathbf{R}}_{\mathbf{c}_j}(n,f)&=\widehat{\mathbf{c}}_j(n,f)\widehat{\mathbf{c}}_j^H(n,f)+(\mathbf{I}-\mathbf{W}_j(n,f))\mathbf{R}_{\mathbf{c}_j}(n,f)
\end{align}
where $\mathbf{R}_{\mathbf{c}_j}(n,f)$ is defined in \eqref{eq:Cov}
and $\mathbf{R}_{\mathbf{x}}(n,f)$ in \eqref{eq:Rx}. In the M-step,
the variance and the spatial covariance of the $j$th source are
updated via
\begin{align}
v_j(n,f)&=\frac{1}{I}\tr(\mathbf{R}_j^{-1}(f)\widehat{\mathbf{R}}_{\mathbf{c}_j}(n,f))
\label{eq:sourcepara}\\
\mathbf{R}_{j}(f)&=\frac{1}{N}\sum_{n=1}^{N}\frac{1}{v_j(n,f)}\widehat{\mathbf{R}}_{\mathbf{c}_j}(n,f)
\label{eq:spatialpara}
\end{align}
where $\tr(.)$ denotes the trace of a square matrix. Note that,
strictly speaking, this algorithm is a generalized form of EM
\cite{McLachlan97}, since the M-step increases but does not maximize
the likelihood of the complete data due to the interleaving of
\eqref{eq:sourcepara} and \eqref{eq:spatialpara}.

\subsection{EM updates for the rank-1 anechoic model and the full-rank direct+diffuse model}
The derivation of EM for the two remaining models is more complex
since the M-step cannot be expressed in closed form. The complete
data and the E-step for the rank-1 anechoic model and the full-rank
direct+diffuse model are identical to those for the rank-1
convolutive model and the full-rank unconstrained model,
respectively. The M-step, which consists of maximizing the
likelihood of the complete data given their natural statistics
computed in the E-step, could be addressed \textit{e.g.} via a
quasi-Newton technique or by sampling possible parameter values from
a grid \cite{Izumi-WASPAA07}. In the following, we do not attempt to
derive the details of these algorithms since these two models appear
to provide lower performance than the rank-1 convolutive model and
the full-rank unconstrained model in a semi-blind context, as
discussed in Section \ref{ssec:exp-semiblind}.

\subsection{Permutation alignment}
\label{ssec:permutation}

Since the parameters of the rank-1 convolutive model and the
full-rank unconstrained model are estimated independently in each
frequency bin $f$, they should be ordered so as to correspond to the
same source across all frequency bins. In order to solve this
so-called permutation problem, we apply the DOA-based algorithm
described in \cite{H.Sawasa-ICASSP06} for the rank-1 model. Given
the geometry of the microphone array, this algorithm computes the
DOAs of all sources and permutes the model parameters by clustering
the estimated mixing vectors $\mathbf{h}_j(f)$ normalized as in
\eqref{eq:normalize}.

\begin{figure}[htb]
\begin{center}
\includegraphics[scale=0.45]{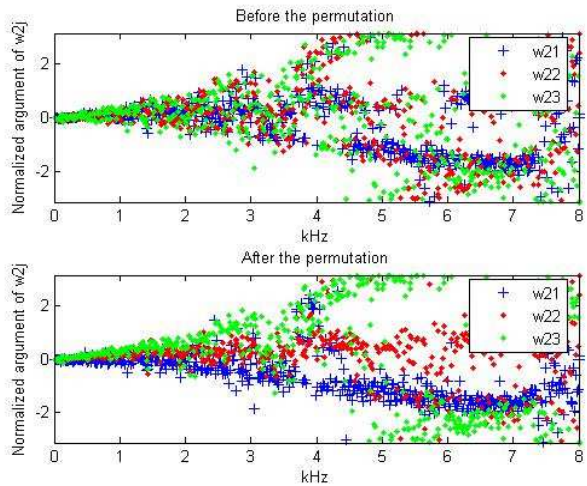}
\caption{Normalized argument of $w_{2j}(f)$ before and after
permutation alignment from a real-world stereo recording of three
sources with $\mathrm{RT}_{60}=250$~ms.} \label{fig:permutation}
\end{center}
\end{figure}

Regarding the full-rank model, we first apply principal component
analysis (PCA) to summarize the spatial covariance matrix
$\mathbf{R}_j(f)$ of each source in each frequency bin by its first
principal component $\mathbf{w}_j(f)$ that points to the direction
of maximum variance. This vector is conceptually equivalent to the
mixing vector $\mathbf{h}_j(f)$ of the rank-1 model. Thus, we can
apply the same procedure to solve the permutation problem. Fig.
\ref{fig:permutation} depicts the phase of the second entry
$w_{2j}(f)$ of $\mathbf{w}_j(f)$ before and after solving the
permutation for a real-world stereo recording of three female speech
sources with room reverberation time $T_{60}=250$~ms, where
$\mathbf{w}_j(f)$ has been normalized as in \eqref{eq:normalize}.
This phase is unambiguously related to the source DOAs below 5 kHz
\cite{H.Sawasa-ICASSP06}. Above that frequency, spatial aliasing
\cite{H.Sawasa-ICASSP06} occurs. Nevertheless, we can see that the
source order is globally aligned for most frequency bins after
solving the permutation.

\section{Experimental evaluation}
\label{sec:experiment} We evaluate the above models and algorithms
under three different experimental settings. Firstly, we compare all
four models in a semi-blind setting so as to estimate an upper bound
of their separation performance. Based on these results, we select
two models for further study, namely the rank-1 convolutive model
and the full-rank unconstrained model. Secondly, we evaluate these
models in a blind setting over synthetic reverberant speech mixtures
and compare them to state-of-the-art algorithms over the real-world
speech mixtures of the 2008 Signal Separation Evaluation Compaign
(SiSEC 2008) \cite{E.Vincent-SiSEC08}. Finally, we assess the
robustness of these two models to source movements in a semi-blind
setting.

\subsection{Common parameter settings and performance criteria}
\label{ssec:expset} The common parameter setting for all experiments
are summarized in Table \ref{table:experimentsetting}. In order to
evaluate the separation performance of the algorithms, we use the
signal-to-distortion ratio (SDR), signal-to-interference ratio
(SIR), signal-to-artifact ratio (SAR) and source image-to-spatial
distortion ratio (ISR) criteria expressed in decibels (dB), as
defined in \cite{E.Vincent-ICA07}. These criteria account
respectively for overall distortion of the target source, residual
crosstalk from other sources, musical noise and spatial or filtering
distortion of the target.

\begin{table}[htb]
\centering
\begin{tabular}{c|c}
\hline \hline
Signal duration & 10 seconds \\
Number of channels & $I=2$ \\
Sampling rate & 16 kHz \\
Window type & sine window \\
STFT frame size & 2048 \\
STFT frame shift & 1024 \\
Propagation velocity & 334 m/s \\
Number of EM iterations & 10 \\
Cluster threshold & $K=30$ \\
\hline \hline
\end{tabular}
\caption{common experimental parameter setting}
\label{table:experimentsetting}
\end{table}

\subsection{Potential source separation performance of all models}
\label{ssec:exp-semiblind} 
\begin{figure}[htb]
\centering
\includegraphics[scale=0.75]{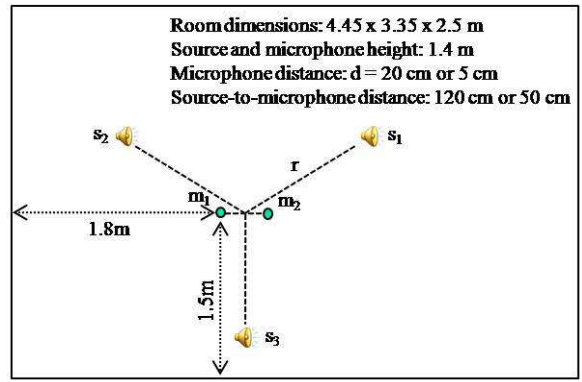}
\caption{Room geometry setting for synthetic convolutive mixtures.}
\label{fig:geometry}
\end{figure}

The first experiment is devoted to the
investigation of the potential source separation performance
achievable by each model in a semi-blind context, \textit{i.e.}
assuming knowledge of the true spatial covariance matrices. We
generated three stereo synthetic mixtures of three speech sources by
convolving different sets of speech signals, \textit{i.e.} male
voices, female voices, and mixed male and female voices, with room
impulse responses simulated via the source image method. The
positions of the sources and the microphones are illustrated in Fig.
\ref{fig:geometry}. The distance from each source to the center of
the microphone pair was 120~cm and the microphone spacing was 20~cm.
The reverberation time was set to $\mathrm{RT}_{60}=250$~ms.

The true spatial covariance matrices $\mathbf{R}_j(f)$ of all
sources were computed either from the positions of the sources and
the microphones and other room parameters or from the mixing
filters. More precisely, we used the equations in Sections
\ref{ssec:model1_conv}, \ref{ssec:model1_ane} and \ref{ssec:model2}
for rank-1 models and the full-rank direct+diffuse model and ML
estimation from the spatial images of the true sources for the
full-rank unconstrained model. The source variances were then
estimated from the mixture using the quasi-Newton technique in
\cite{Duong-WASPAA09}, for which an efficient initialization exists
when the spatial covariance matrices are fixed. Binary masking and
$\ell_1$-norm minimization were also evaluated for comparison using
the same mixing vectors as the rank-1 convolutive model with the
reference software in \cite{E.Vincent-SiSEC08}. The results are
averaged over all sources and all set of mixtures and shown in Table
\ref{table:potentialSxR}.

\begin{table}[htb]
\centering
\begin{tabular}{|c|c|cccc|}
\hline
\begin{minipage}{1.3cm}\centering\vspace{1pt} Covariance\\models\vspace{1pt}\end{minipage} & \begin{minipage}{1.3cm}\centering\vspace{1pt} Number of spatial parameters\vspace{1pt}\end{minipage} & SDR & SIR & SAR & ISR \\
%Covariance model & Number of spatial parameters & SDR & SIR & SAR & ISR\\
\hline
Rank-1 anechoic & 6 & 0.8 & 2.4 & 7.9 & 5.0\\
Rank-1 convolutive & 3078 & 3.8 & 7.5 & 5.3 & 9.3\\
Full-rank direct+diffuse & 8 & 3.2 & 6.9 & 5.4 & 7.9\\
Full-rank unconstrained & 6156 & 5.6 & 10.7 & 7.3 & 11.0\\
Binary masking & 3078 & 3.3 & 11.1 & 2.4 & 8.4\\
$\ell_1$-norm minimization & 3078 & 2.7 & 7.7 & 3.4 & 8.6\\
\hline
\end{tabular}
\caption{Average potential source separation performance in a
semi-blind setting over stereo mixtures of three sources with
$\mathrm{RT}_{60}=250$~ms.} \label{table:potentialSxR}
\end{table}

The rank-1 anechoic model has lowest performance because it only
accounts for the direct path. By contrast, the full-rank
unconstrained model has highest performance and improves the SDR by
1.8~dB, 2.3~dB, and 2.9~dB when compared to the rank-1 convolutive
model, binary masking, and $\ell_1$-norm minimization respectively.
The full-rank direct+diffuse model results in a SDR decrease of
0.6~dB compared to the rank-1 convolutive model. This decrease
appears surprisingly small when considering the fact that the former
involves only 8 spatial parameters (6 distances $r_{ij}$, plus
$\sigma_\mathrm{rev}^2$ and $d$) instead of 3078 parameters (6
mixing coefficients per frequency bin) for the latter. Nevertheless,
we focus on the two best models, namely the rank-1 convolutive model
and the full-rank unconstrained model in subsequent experiments.

\subsection{Blind source separation performance  as a function of the reverberation time}
\label{ssec:blind-T60change} 
\begin{figure*}[t]
\begin{center}
\includegraphics[scale=0.3]{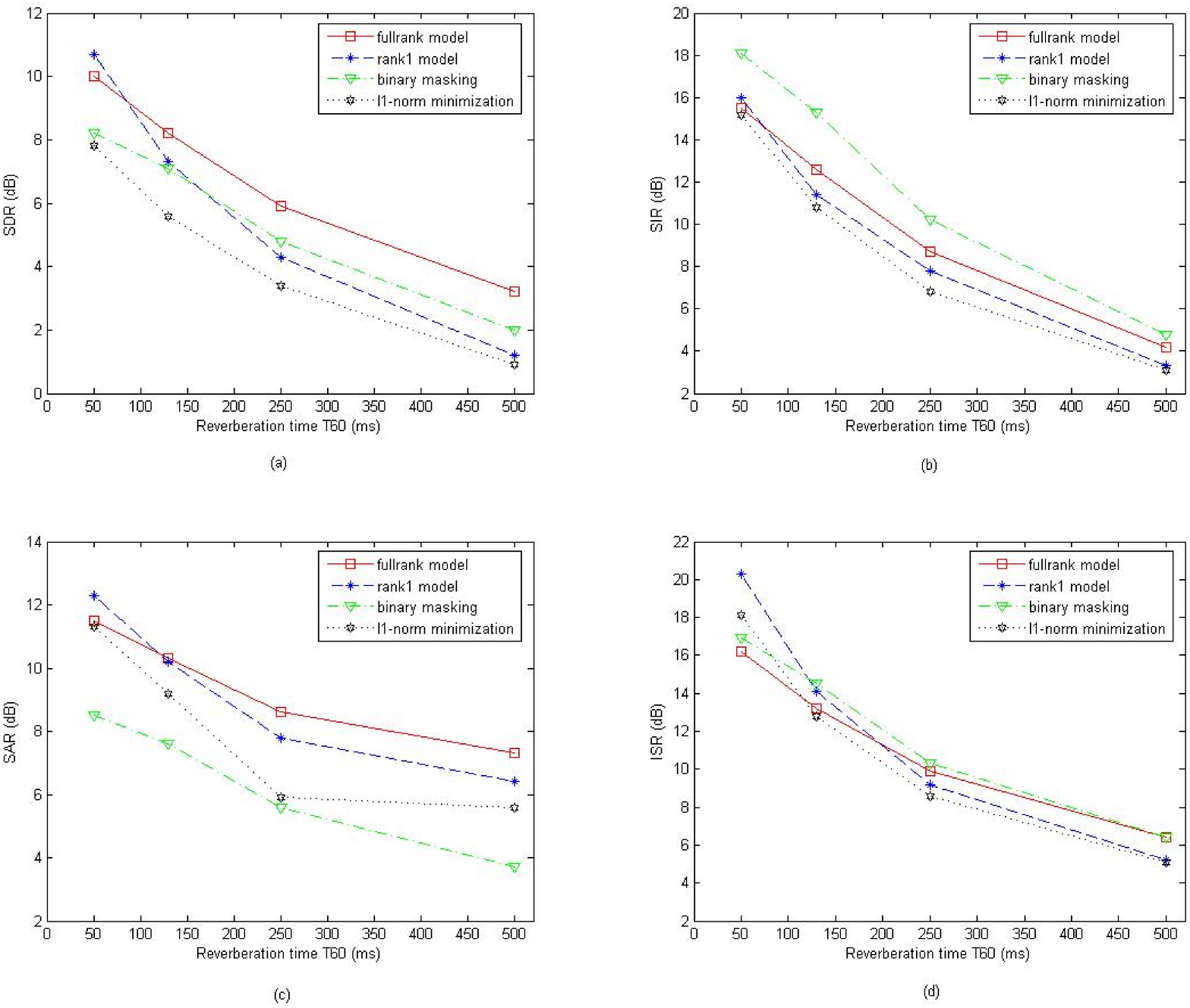}
\end{center}
\caption{Average blind source separation performance over stereo
mixtures of three sources as a function of the reverberation time.}
\label{fig:SxRT60change}
\end{figure*}

The second experiment aims to
investigate the blind source separation performance achieved via
these two models and via binary masking and $\ell_1$-norm
minimization in different reverberant conditions. Synthetic speech
mixtures were generated in the same as in the first experiment,
except that the microphone spacing was changed to 5~cm and the
distance from the sources to the microphones to 50~cm. The
reverberation time was varied in the range from 50 to 500~ms. The
resulting source separation performance in terms of SDR, SIR, SAR,
and ISR is depicted in Fig. \ref{fig:SxRT60change}.

We observe that in a low reverberant environment, \textit{i.e.}
$T_{60}=50$~ms, the rank-1 convolutive model provides the best SDR
and SAR. This is consistent with the fact that the direct part
contains most of the energy received at the microphones, so that the
rank-1 spatial covariance matrix provides similar modeling accuracy
than the full-rank model with fewer parameters. However, in an
environment with realistic reverberation time, \textit{i.e.}
$T_{60}\ge130$~ms, the full-rank unconstrained model outperforms
both the rank-1 model and binary masking in terms of SDR and SAR and
results in a SIR very close to that of binary masking. For instance,
with $T_{60}=500$~ms, the SDR achieved via the full-rank
unconstrained model is $2.0$~dB, $1.2$~dB and $2.3$~dB larger than
that of the rank-1 convolutive model, binary masking, and
$\ell_1$-norm minimization respectively. These results confirm the
effectiveness of our proposed model parameter estimation scheme and
also show that full-rank spatial covariance matrices better
approximate the mixing process in a reverberant room.

\subsection{Blind source separation with the SiSEC 2008 test data}
\label{ssec:compareSiSEC08} We conducted a third experiment to
compare the proposed full-rank unconstrained model-based algorithm
with state-of-the-art BSS algorithms submitted for evaluation to
SiSEC 2008 over real-world mixtures of 3 or 4 speech sources. Two
mixtures were recorded for each given number of sources, using
either male or female speech signals. The room reverberation time
was 250~ms and the microphone spacing 5~cm \cite{E.Vincent-SiSEC08}.
The average SDR achieved by each algorithm is listed in Table
\ref{table:SiSECcompare}. The SDR figures of all algorithms except
yours were taken from the website of SiSEC
2008\footnote{http://sisec2008.wiki.irisa.fr/tiki-index.php?page=Under-determined+speech+and+music+mixtures}.

\begin{table}[htb]
\centering
\begin{tabular}{|c|c|c|}
\hline
Algorithms & 3 source mixtures & 4 source mixtures \\
\hline
full-rank unconstrained & 3.8 & 2.0\\
M. Cobos \cite{Cobos-IEEETrans09} & 2.2 & 1.0\\
M. Mandel \cite{Mandel-WASPAA07} & 0.8 & 1.0\\
R. Weiss \cite{Weiss-10} & 2.3 & 1.5\\
S. Araki \cite{Araki-ICA09} & 3.7 & - \\
Z. El Chami \cite{Chami-IWAENC08} & 3.1 & 1.4\\
\hline
\end{tabular}
\caption{Average SDR over the real-world test data of SiSEC 2008
with $T_{60}=250$~ms and 5~cm microphone spacing.}
\label{table:SiSECcompare}
\end{table}

For three-source mixtures, our algorithm provides 0.1 dB SDR
improvement compared to the best current result given by Araki's
algorithm \cite{Araki-ICA09} . For four-source mixtures, it provides
even higher SDR improvement of 0.5~dB compared to the best current
result given by Weiss's algorithm \cite{Weiss-10}.

\subsection{Investigation of the robustness to small source movements}
\label{ssec:DoAchanges} 
\begin{figure}[h]
\begin{minipage}[t]{2.7in}
\begin{center}
\includegraphics[scale=0.45]{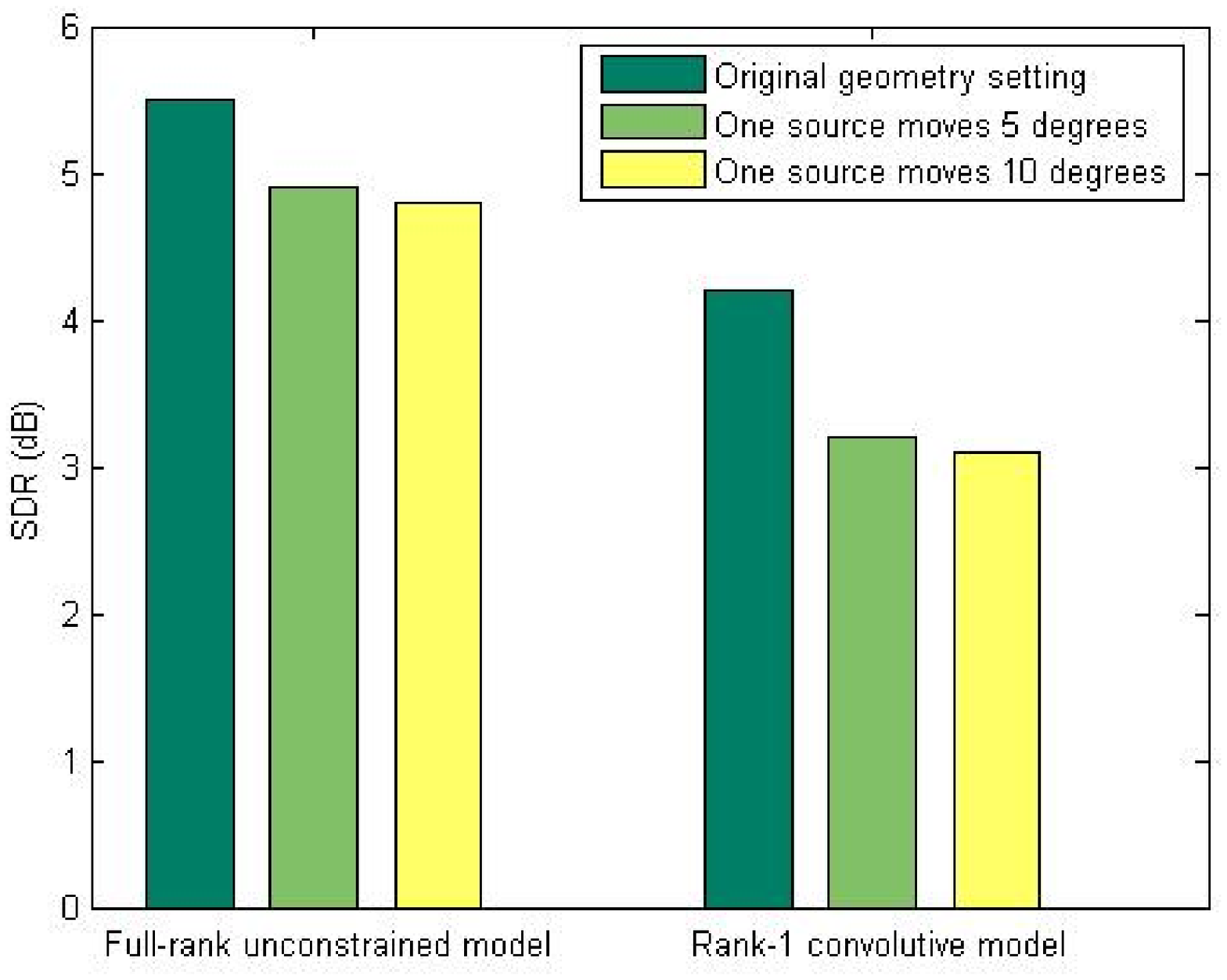}
\end{center}
\caption{SDR results in the small source movement scenarios.}
\label{fig:SDR_DoAchange}
\end{minipage}
\vfill
\begin{minipage}[t]{2.7in}
\begin{center}
\includegraphics[scale=0.45]{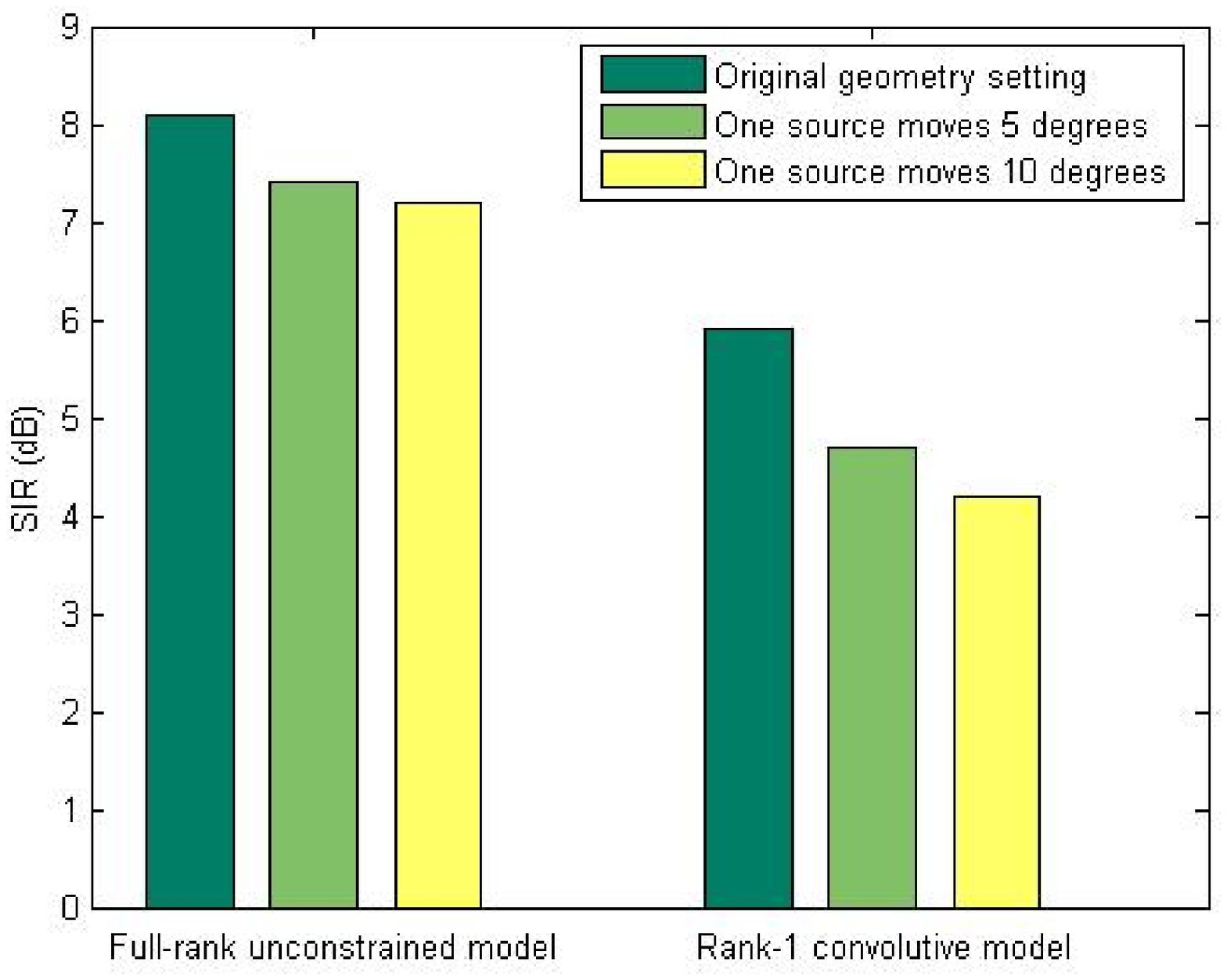}
\end{center}
\caption{SIR results in the small source movement scenarios.}
\label{fig:SIR_DoAchange}
\end{minipage}
\end{figure}

Our last experiment aims to to examine the
robustness of the rank-1 convolutive model and the full-rank
unconstrained model to small source movements. We made several
recordings of three speech sources $s_1$, $s_2$, $s_3$ in a meeting
room with 250~ms reverberation time using omnidirectional
microphones spaced by 5~cm. The distance from the sources to the
microphones was 50~cm. For each recording, the spatial images of all
sources were separately recorded and then added together to obtain a
test mixture. After the first recording, we kept the same positions
for $s_1$ and $s_2$ and successively moved $s_3$ by 5 and 10$^\circ$
both clock-wise and counter clock-wise resulting in 4 new positions
of $s_3$. We then applied the same procedure to $s_2$ while the
positions of $s_1$ and $s_3$ remained identical to those in the
first recording. Overall, we collected nine mixtures: one from the
first recording, four mixtures with 5$^\circ$ movement of either
$s_2$ or $s_3$, and four mixtures with 10$^\circ$ movement of either
$s_2$ or $s_3$. We performed source separation in a semi-blind
setting: the source spatial covariance matrices were estimated from
the spatial images of all sources recorded in the first recording
while the source variances were estimated from the nine mixtures
using the same algorithm as in Section \ref{ssec:exp-semiblind}. The
average SDR and SIR obtained for the first mixture and for the
mixtures with 5$^\circ$ and 10$^\circ$ source movement are depicted
in Fig. \ref{fig:SDR_DoAchange} and Fig. \ref{fig:SIR_DoAchange},
respectively. This procedure simulates errors encountered by on-line
source separation algorithms in moving source environments, where
the source separation parameters learnt at a given time are not
applicable anymore at a later time.

The separation performance of the rank-1 convolutive model degrades
more than that of the full-rank unconstrained model both with
5$^\circ$ and 10$^\circ$ source rotation. For instance, the SDR
drops by 0.6~dB for the full-rank unconstrained model based
algorithm when a source moves by 5$^\circ$ while the corresponding
drop for the rank-1 convolutive model equals 1~dB. This result can
be explained when considering the fact that the full-rank model
accounts for the spatial spread of each source as well as its
spatial direction. Therefore, small source movements remaining in
the range of the spatial spread do not affect much separation
performance. This result indicates that, besides its numerous
advantages presented in the previous experiments, this model could
also offer a promising approach to the separation of moving sources
due to its greater robustness to parameter estimation errors.

\section{Conclusion and discussion}
\label{sec:conclusion}

In this article, we presented a general probabilistic framework for
convolutive source separation based on the notion of spatial
covariance matrix. We proposed four specific models, including
rank-1 models based on the narrowband approximation and full-rank
models that overcome this approximation, and derived an efficient
algorithm to estimate their parameters from the mixture.
Experimental results indicate that the proposed full-rank
unconstrained spatial covariance model better accounts for
reverberation and therefore improves separation performance compared
to rank-1 models and state-of-the-art algorithms in realistic
reverberant environments.

Let us now mention several further research directions. Short-term
work will be dedicated to the modeling and separation of diffuse and
semi-diffuse sources or background noise via the full-rank
unconstrained model. Contrary to the rank-1 model in
\cite{Ozerov-IEEETrans10} which involves an explicit spatially
uncorrelated noise component, this model implicitly represents noise
as any other source and can account for multiple noise sources as
well as spatially correlated noises with various spatial spreads. A
further goal is to complete the probabilistic framework by defining
a prior distribution for the model parameters across all frequency
bins so as to improve the robustness of parameter estimation with
small amounts of data and to address the permutation problem in a
probabilistically relevant fashion. Finally, a promising way to
improve source separation performance is to combine the spatial
covariance models investigated in this article with models of the
source spectra such as Gaussian mixture models
\cite{S.Arberet-ICA09} or nonnegative matrix factorization
\cite{Ozerov-IEEETrans10}.

\bibliographystyle{IEEEtran}
\bibliography{AllBSS}

\end{document}